\documentclass[10pt,twocolumn,letterpaper]{article}

\usepackage{iccv}
\usepackage{times}
\usepackage{epsfig}
\usepackage{graphicx}
\usepackage{amsmath}
\usepackage{amssymb}

\usepackage[pagebackref=true,breaklinks=true,letterpaper=true,colorlinks,bookmarks=false]{hyperref}

 \iccvfinalcopy 

\def\iccvPaperID{680} 

\ificcvfinal\fi
\begin{document}

\title{Fusing Multi-Stream Deep Networks for Video Classification}

\author{Zuxuan Wu, Yu-Gang Jiang, Hao Ye, Xi Wang, Xiangyang Xue\\             
Fudan University\\
{\tt\small zxwu, ygj, haoye10, xwang10, xyxue@fudan.edu.cn}
\and
Jun Wang\\
East China Normal University\\
{\tt\small wongjun@gmail.com}
}
\maketitle


\begin{abstract}
This paper studies deep network architectures to address the problem of video classification. A multi-stream framework is proposed to fully utilize the rich multimodal information in videos. Specifically, we first train three Convolutional Neural Networks to model spatial, short-term motion and audio clues respectively. Long Short Term Memory networks are then adopted to explore long-term temporal dynamics. With the outputs of the individual streams, we propose a simple and effective fusion method to generate the final predictions, where the optimal fusion weights are learned adaptively for each class, and the learning process is regularized by automatically estimated class relationships. Our contributions are two-fold. First, the proposed multi-stream framework is able to exploit multimodal features that are more comprehensive than those previously attempted. Second, we demonstrate that the adaptive fusion method using the class relationship as a regularizer outperforms traditional alternatives that estimate the weights in a ``free"  fashion. Our framework produces significantly better results than the state of the arts on two popular benchmarks, 92.2\% on UCF-101 (without using audio) and 84.9\% on Columbia Consumer Videos.
\end{abstract}

\section{Introduction}
The problem of video classification based on semantic contents like human actions or complex events has been extensively studied in the computer vision community. The fact that videos are intrinsically multimodal demands solutions that can explore not only static visual information, but also motion and auditory clues. Key to the development of video classification systems is the design of good features. Popular feature descriptors include the SIFT \cite{DBLP:journals/ijcv/Lowe04}, the Mel-Frequency Cepstral Coefficients (MFCC) \cite{xu2013feature}, the STIP \cite{laptevSTIP} and the dense trajectories~\cite{wang2013action}, which can be encoded into video-level representations by bag-of-words (BoW) \cite{sun2009action,ye2012robust,natarajan2012multimodal} or Fisher vectors (FV) \cite{oneata2013action,sanchez2013image,lan2014beyond,zha2015exploiting}.

\begin{figure}[t!]
\centering
\epsfig{file=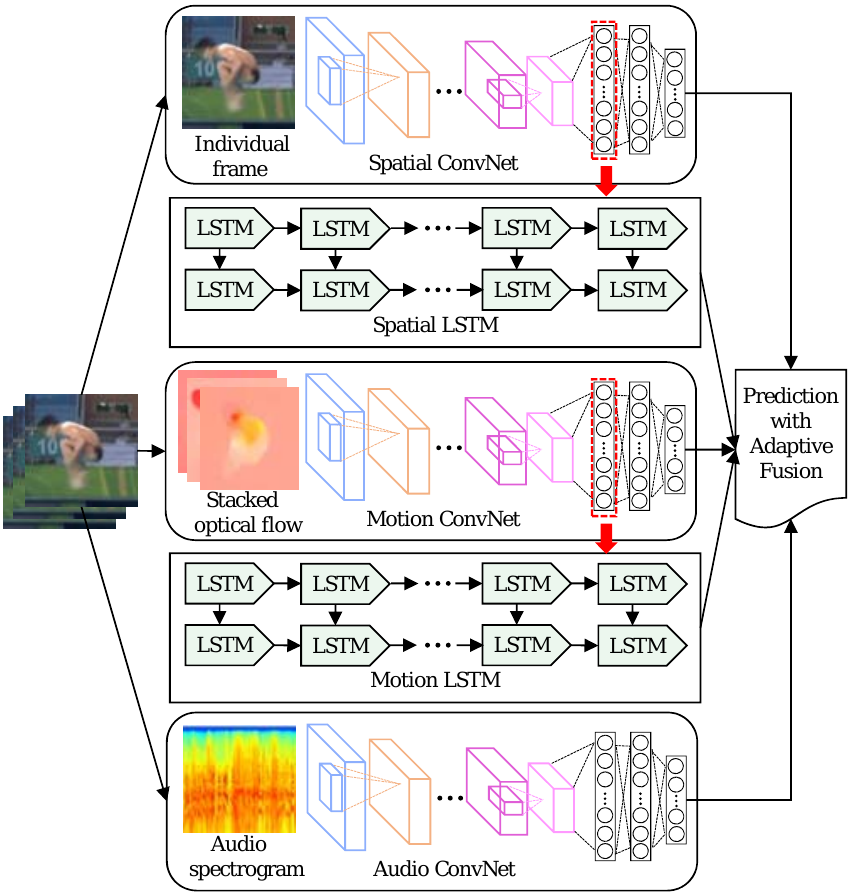, scale=0.89}
\caption{Illustration of the proposed multi-stream video classification framework.}
\label{fig:framework}
\end{figure}
\vspace{-1em}

In contrast to the hand-engineered descriptors, the deep neural networks that can learn features automatically from raw data have demonstrated strong performance in various domains. In particular, the convolutional neural networks (ConvNets) are very successful on image analysis tasks like object detection~\cite{girshick2014rcnn}, object recognition~\cite{simonyan2014very,Szegedy:2014tb} and image segmentation~\cite{farabet2013learning}. However, for video classification, most deep network based approaches (\eg,~\cite{DBLP:conf/icml/JiXYY10,KarpathyCVPR14,DBLP:conf/nips/SimonyanZ14}) demonstrated worse or similar results to the hand-engineered features~\cite{wang2013action}. This is largely due to the high complexity of the video data. Unlike images that only have static visual appearance information, videos also contain temporal motions and auditory soundtracks. For example, a ``diving'' action video usually involves a sequence of atoms, such as ``jumping from a platform'', ``rotating in the air'' and ``falling into water'', accompanied by cheering or clapping sounds. Some approaches~\cite{DBLP:conf/icml/JiXYY10,KarpathyCVPR14,DBLP:conf/nips/SimonyanZ14} only  focused on the static frames and short-term motion clues captured by a few adjacent frames, which are apparently not sufficient. A few very recent studies attempted to use recurrent neural networks (RNN) to model long-term temporal information and achieved competitive performance~\cite{ng2015beyond,wu2015modeling}. Nevertheless, the audio information has rarely been exploited. In addition, most existing approaches fused the outputs of multiple networks in a very straightforward way~\cite{DBLP:conf/nips/SimonyanZ14}, which could lead to sub-optimal performance.

Realizing the above limitations, in this paper, we propose a multi-stream framework of deep neural networks to exploit the multimodal clues for video classification. Figure~\ref{fig:framework} illustrates the diagram of our approach. Three ConvNets are trained to model the static spatial information, short-term motion and auditory clues, respectively. The motion stream is computed on stacked optical flows over a short temporal windows and thus can only capture short-term motion.  In order to model the long-term temporal clues, we employ a Recurrent Neural Network (RNN) model, namely the Long Short Term Memory (LSTM), on the frame-level spatial and motion features extracted by the ConvNets. The LSTM encodes history information in memory units regulated with non-linear gates to discover temporal dependencies. To combine the outputs from different networks, we develop a simple yet effective fusion method to learn the optimal fusion weights adaptively for each class. We propose to regularize the weight learning process using class relationships estimated without using additional labels. This helps inject class context into the final predictions and thus can significantly improve the results. Our contributions are summarized as follows:
\begin{enumerate}
\item We introduce a multi-stream framework that integrates spatial, short-term motion, long-term temporal and auditory clues in videos. We demonstrate that the multi-stream networks are able digest complementary information to receive significantly improved performance. 
\item We propose a simple and effective fusion method to combine the outputs of the individual networks. The method learns fusion weights adaptively for each class and is able to harness class relationships in the weight learning process. We empirically show that the class relationship regularizer is very effective. 
\end{enumerate}
Incorporating the fusion method into the proposed multi-stream framework, we achieve superior performance on two popular benchmark datasets. 

\section{Related Works}
As aforementioned, video classification has been extensively studied and significant efforts have been paid to design hand-engineered features or classifiers. We focus the review on recent works related to our proposed approach.

Motived by the promising results of deep networks (particularly the ConvNets) on image analysis tasks~\cite{Szegedy:2014tb,simonyan2014very,girshick2014rcnn}, several works have exploited deep architectures for video classification. Ji \etal extended CNN models into spatial-temporal space by operating on stacked video frames~\cite{DBLP:conf/icml/JiXYY10}.  Karparthy \etal compared several architectures for action recognition~\cite{KarpathyCVPR14}. Tran~\etal proposed to learn generic spatial-temporal features which can be computed efficiently~\cite{tran2014c3d}. Simonyan and Zisserman~\cite{DBLP:conf/nips/SimonyanZ14} introduced an interesting two-stream approach, where two ConvNets are trained to explicitly capture spatial and short-term motion information using frames and stacked optical flows as inputs, respectively. Final predictions can be obtained by linearly averaging the prediction scores of the two ConvNets. In this paper, we also adopt two similar ConvNets as~\cite{DBLP:conf/nips/SimonyanZ14}. However, as the two-stream approach is not able to model the auditory and the long-term temporal clues, we adopt additional networks to build a more comprehensive framework. A novel fusion method is also proposed to combine the multi-stream outputs, which is better than the simple linear fusion used in~\cite{DBLP:conf/nips/SimonyanZ14}.

The RNN has been shown to be effective on many sequential modeling tasks, such as speech recognition~\cite{DBLP:conf/icassp/GravesMH13} and image/video description~\cite{DBLP:journals/corr/DonahueHGRVSD14,yao2015video}. For long-term temporal modeling of the video data, Srivastava \etal proposed an LSTM encoder-decoder framework to learn video representations in an unsupervised manner~\cite{DBLP:journals/corr/SrivastavaMS15}. Donahua \etal \cite{DBLP:journals/corr/DonahueHGRVSD14} and Wu \etal \cite{wu2015modeling} trained a two-layer LSTM network for action classification. Ng \etal~\cite{ng2015beyond} further demonstrated that a five-layer LSTM network is slightly better.

Fusion is needed to combine the outputs of separate prediction models. The simplest solution is linear weighted fusion, which has been adopted in many recent approaches like~\cite{DBLP:conf/nips/SimonyanZ14}. Nandakumar~\etal performed score fusion using a method called likelihood ratio test~\cite{nandakumar2008likelihood}. More recently, Xu \etal~\cite{xu2013feature} and Ye \etal~\cite{ye2012robust} proposed robust late fusion methods by seeking a low rank matrix to remove the noise of individually trained classifiers. Liu et al.~\cite{liu2013sample} proposed to predict sample-specific weights in the fusion process. 

There are many studies using context or class relationships to improve visual recognition performance. For instance, Rabinovich~\etal utilized a Conditional Random Field (CRF) model to maximize object label agreement based on contextual relevance~\cite{rabinovich2007objects}. Deng~\etal proposed to jointly train a hierarchy and exclusion graph model with a ConvNet to learn class relations for image classification~\cite{DengJiaECCV2014}. Assari~\etal exploited class co-occurrences for video classification~\cite{AssariUCF-CVPR2014}. Different from these works, we use class relationship as a regularizer to learn fusion weights adaptively for each class.


\section{Methodology}
In this section, we first describe the individual streams and then introduce the proposed adaptive multi-stream fusion method, followed by implementation details.

\subsection{Multi-Stream ConvNets}
Carrying abundant multimodal information, videos normally show the movements and interactions of objects under certain scenes over time, accompanied by human voices or background sounds. Therefore, video data can be naturally decomposed into spatial, motion and audio streams. The spatial stream consisting of individual frames depicts the static appearance information, while the motion stream captures object or scene movements demonstrated by continuous frames. In addition, sounds in the audio stream provide crucial clues that are often complementary to the visual counterpart. Motivated by the recent two-stream approach~\cite{DBLP:conf/nips/SimonyanZ14}, we train three ConvNets to exploit the multimodal information, as described below.

In brief, the spatial ConvNet uses the raw frames as inputs, where we adopt a deep architecture with superior performance on image recognition tasks~\cite{simonyan2014very}. It can effectively recognize certain video semantics that have clear and discriminative appearance characteristics. For the motion stream, we train a ConvNet model operating on stacked optical flows following~\cite{DBLP:conf/nips/SimonyanZ14}. More specifically, through computing displacement vectors in both horizontal and vertical ways, the optical flows encode subtle motion patterns of objects between each pair of adjacent frames, which can be converted into two flow images as the inputs of the motion stream ConvNet. Previous studies have shown that further improvements can be obtained by stacking consecutive optical flow images in a short time window, owing to the inclusion of relatively more compact movements~\cite{DBLP:conf/nips/SimonyanZ14}. In order to leverage the audio information, we first apply the Short-Time Fourier Transformation to convert the 1-d soundtrack into a 2-D image (namely spectrogram) with the horizontal axis and vertical axis being time-scale and frequency-scale respectively. Then we employ a ConvNet to operate on the spectrograms as suggested in~\cite{van2013deep}. Notice that the ConvNet is well suited for modeling audio signals based on spectrograms with the weight sharing and max pooling mechanism to strive invariance of small frequency shifts~\cite{abdel2013exploring}. 

\subsection{Long Term Temporal Modeling} 
As the motion stream ConvNet only captures short-term motion patterns, we further employ LSTM~\cite{hochreiter1997long} to model long-term temporal clues in the visual channel. LSTM is a popular RNN model that incorporates memory cells with several gates to learn long-term dependencies without suffering from vanishing and exploding gradients as the traditional RNNs~\cite{bengio1994learning}. It is able to exploit temporal information of a data sequence with arbitrary length through recursively mapping the input sequence to output labels with hidden LSTM units. Each of the units maintains a built-in memory cell, which stores information over time guarded by several non-linear gate units to control the amount of changes and influence of the memory contents.

\begin{figure}[t!]
\centering
\epsfig{file=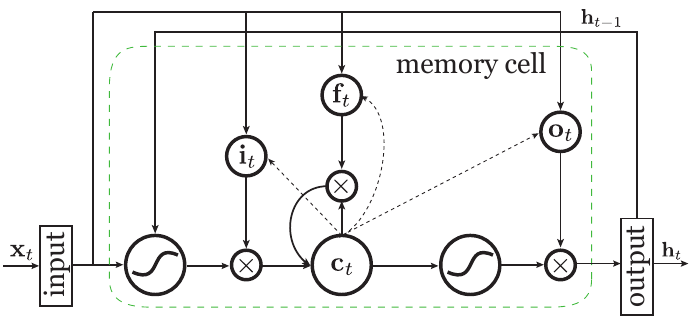, scale=1.1}
\caption{\label{fig:lstmunit}The structure of an LSTM unit.}
\end{figure}

Figure~\ref{fig:lstmunit} illustrates the typical structure of a hidden LSTM unit. In our framework, we denote ${\bf x}_t$ as the feature representation of a video frame or a stacked optical flow image at the $t$-th time step. Generally, an LSTM maps an input sequence $({\bf x}_1,{\bf x}_2,\ldots,{\bf x}_T)$ to output labels $({\bf y}_1,{\bf y}_2,\ldots,{\bf y}_T)$ through computing activations of the units in the network recursively from $t=1$ to $t = T$. At time~$t$, the activation vectors of memory cell ${\bf c}_t$, output gate ${\bf o}_t$ and hidden state ${\bf h}_t$ are computed as:
\begin{align} \nonumber
& {\bf c}_t={\bf f}_t\odot{\bf c}_{t-1}+{\bf i}_t\odot\tanh({\bf W}_{xc}{\bf x}_t+{\bf W}_{hc}{\bf h}_{t-1}+{\bf b}_c), \\ \nonumber
& {\bf o}_t=\sigma({\bf W}_{xo}{\bf x}_t+{\bf W}_{ho}{\bf h}_{t-1}+{\bf W}_{co}{\bf c}_{t}+{\bf b}_o),\\ 
& {\bf h}_t={\bf o}_t\odot\tanh({\bf c}_t),
\end{align}
where ${\bf W}_{xc}, {\bf W}_{hc}, {\bf W}_{xo}, {\bf W}_{ho}, {\bf W}_{co}$ are the weight matrices connecting two different units.  ${\bf b}_c, {\bf b}_o$ are the bias terms, $\sigma$ is the sigmoid function, and $\odot$ is an element-wise product operator. Notice that ${\bf i}_t$ and ${\bf f}_t$ are the activation vectors of input and forget gates, which are calculated with weight matrices as:
\begin{align} \nonumber
& {\bf i}_t=\sigma({\bf W}_{xi}{\bf x}_t+{\bf W}_{hi}{\bf h}_{t-1}+{\bf W}_{ci}{\bf c}_{t-1}+{\bf b}_i), \\
& {\bf f}_t=\sigma({\bf W}_{xf}{\bf x}_t+{\bf W}_{hf}{\bf h}_{t-1}+{\bf W}_{cf}{\bf c}_{t-1}+{\bf b}_f).
\end{align}

From the above equations, the contents of the memory cell at the $t$-th time step ${\bf c}_t$ is computed as the weighted sum of the current inputs and the previous memory contents ${\bf c}_{t-1}$. The input and forget gates (\textit{i.e.}, ${\bf i}_t$ and ${\bf f}_t$) impose regularization to determine whether to consider new information or forget old information. In addition, the output gate ${\bf o}_t$ controls the amount of information from the memory contents that is passed to the hidden state ${\bf h}_t$ to influence the computation in the next time step. 

As a neural network, the LSTM model can be easily deepened by stacking the hidden states from a layer $l-1$ as inputs of the next layer $l$.  In order to obtain the prediction scores for a total of $C$ classes at a time step $t$, a softmax layer is placed on top of the last LSTM layer $L$ to estimate the posterior probability $p_c$ of the $c$-th class as:
\begin{align}
& p_c~=~\text{softmax}({\bf h}^L_t)~=~~\frac{\exp({{\bf u}_c}^T{\bf h}^L_{t}+b_c)}{\sum_{c'\in C}\exp({{\bf u}_{c'}}^T{\bf h}^L_{t}+b_{c'})},
\end{align}
where ${\bf u}_c$ and $b_c$ represent the corresponding weight vector and the bias term of the $c$-th class. Such an LSTM network can be trained using the Back-Propagation Through Time (BPTT) algorithm~\cite{graves2005framewise},  which ``unrolls'' the model into a feed forward neural net and back-propagates to determine the optimal network parameters. We adopt the output from the last layer as the video-level prediction scores since this output is computed based on the information from the entire sequence. Our empirical results show that using the last layer output is better than pooling the predictions at all the time steps.

\subsection{Adaptive Multi-Stream Fusion}
Given the prediction scores of multiple network streams (\ie, the ConvNets and the LSTM), we are able to capture the video characteristics from different aspects. It is critical to effectively fuse the multi-stream scores to generate the final predictions. Different semantic classes associate with the multiple streams with different strength. For example, some classes are strongly associated with particular objects which could be effectively recognized with the spatial stream, while others may contain dramatic movements so the short-term motion and the long-term temporal clues can contribute more significantly. Traditional fusion methods are usually performed at the stream-level without considering  the class-specific preferences. In addition, most existing studies on model fusion neglected the class relationships that can serve as complementary information for improved performance~\cite{rabinovich2007objects,bengio2013using,DengJiaECCV2014}. In the following we introduce the proposed adaptive multi-stream fusion method, which is able to determine the optimal fusion weights adaptively for each class. The highly correlated classes are also automatically identified and their relationships are utilized in the method. 

Formally, we denote the prediction scores from the $m$-th stream as ${\bf s}^m \in \mathbb{R}^C$ ($m = 1, \cdots, M$) with $C$ being the number of classes, and let ${\bf \hat{y}}$ be the final predicted labels. A straightforward way of  late fusion is to compute the final prediction as ${\bf \hat{y}}  = f({\bf s}^1, \cdots, {\bf s}^M)$. Here $f$ is a transition function, which can be a linear function, a logistic function, \etc. However, such a late fusion approach treats all the classes uniformly without considering their different characteristics. 

Different from the uniform fusion methods, we attempt to adaptively integrate the predictions from multiple streams for each class by not only combining scores across streams but also utilizing class knowledge as a prior to provide additional information. To this end, we first stack the multiple score vectors of a training sample $n$ as a coefficient vector: $ {\bf s}_n~=~\left[ {{\bf s}_n^1}^\top, \cdots, {{\bf s}_n^m}^\top, \cdots, {{\bf s}_n^M}^\top \right]^\top  \in \mathbb{R}^{CM} $. Then the best class-specific fusion weights can be learned with logistic regression as:
\begin{align}
{\bf W}  = &\arg\min_{{\bf w}, \cdots, {\bf w}_C} \sum_{n,c} \text{log}\left (1+\text{exp}\left [(1-2y_{n,c}){\bf s}_n^T{\bf w}_c\right ]\right ),
\label{eq:logistic}
\end{align}
where $y_{n,c}$ is the ground-truth label of the $n$-th sample for class $c$, and ${\bf W} = \left[{\bf w}_1, \cdots, {\bf w}_c, \cdots, {\bf w}_C\right] \in \mathbb{R}^{CM \times C}$. However, direct optimization with the above formulation often leads to over-fitting and produces limited performance on the test set. In order to alleviate this and take the class relationships into account, we use the relationships as a prior to guide the learning of the weights. More precisely, we first compute a correlation matrix ${\bf V}^m \in \mathbb{R}^{C \times C}$ of the classes for the $m$-th stream using the corresponding prediction scores, where each entry ${\bf V}_{ij}$ indicates the percentage of the samples with the ground-truth label of class~$i$ being wrongly classified into class $j$.  The reason of using separate correlation matrix for each stream is that the captured class relationships in different streams are likely to be quite different. Next, we stack the similarity matrices of all the streams ${\bf V} = \left[ {\bf V}^1, \cdots, {\bf V}^m, \cdots, {\bf V}^M \right]^\top$ to regularize the weight learning process as:
\begin{equation}
\label{eq:obj1}
    \min_{{\bf W}} ~~L({\bf S}, {\bf Y}; {\bf W}) + \lambda_1 \left\|{\bf W} - {\bf V} \right\|_F^2, 
\end{equation}
where the first term is the empirical loss that measures the discrepancy between the ground-truth labels ${\bf Y}$ and the prediction scores ${\bf S}$, and the second term regularizes the fusion weights using the class correlation as a prior. 
For each similarity matrix ${\bf V}^m$, the non-diagonal entries demonstrate the similarities among different classes, which can be used to guide the weight learning process through borrowing information from highly related classes. 

In addition, we also incorporate an $\ell_1$ norm regularization to impose sparsity on the weight matrix, which, to some extent, can help avoid information sharing from irrelevant classes. With both regularization terms, we have following optimization problem:
\begin{equation}
\label{eq:finalobj}
    \min_{{\bf W}} ~~L({\bf S}, {\bf Y}; {\bf W}) + \lambda_1 \left\|{\bf W} - {\bf V} \right\|_F^2 +  \lambda_2 \left\|{\bf W} \right\|_1.
\end{equation}
In summary, by treating the class correlation matrix as a prior, our fusion approach minimizes an empirical loss regularized by a sparsity constraint to effectively derive class adaptive fusion weights.

Although the loss function in Equation~\ref{eq:finalobj} is convex, it is non-trivial to solve it due to the non-smooth term. To tackle the optimization problem efficiently, we adopt the proximal gradient descent method that splits the objective function into a smooth part and a non-smooth part:
\begin{align}
            g &= L({\bf S}, {\bf Y}; {\bf W}) + \lambda_1 \left\|{\bf W} - {\bf V} \right\|_F^2, \\
            h &= \lambda_2 \left\|{\bf W} \right\|_1.
\end{align}
The update of ${\bf W}$ at the $k+1$ iteration can be simply computed as: 
\begin{equation*}
    {\bf W}^{k+1} = \text{Prox}_h ({\bf W}^{k} - \nabla g({\bf W}^{k})),
\end{equation*}
 where $\text{Prox}_h$ denotes the soft-thresholding operator for the $\ell_1$ norm~\cite{donoho1995adapting}. 

Note that the additional computational cost lies in the estimation of the proximal operator. Since it can be analytically solved in linear time~\cite{bach2011convex}, the above optimization process is fairly efficient.

\subsection{Implementation Details and Discussions}
{\textbf{ConvNet Models}}. In this work, we adopt two ConvNet architectures, the CNN\_M~\cite{DBLP:conf/nips/SimonyanZ14} model for capturing the short-term motion and the audio clues and a recent deeper VGG\_19~\cite{simonyan2014very} architecture for the spatial stream. The CNN\_M is basically a variant of the AlexNet~\cite{DBLP:conf/nips/KrizhevskySH12} with more filters included, which contains five convolutional layers followed by three fully connected layers.  The VGG\_19 not only reduces the size of the convolutional filters and the stride, but also extends the depth of the network to a total of 19 layers, equipping the architecture with the capacity of learning more robust representations. These two deep networks achieved 13.5\%~\cite{DBLP:conf/nips/SimonyanZ14} and 7.5\%~\cite{simonyan2014very} top-5 error rates on the ImageNet ILSVRC-2012 validation set, respectively. All the ConvNet models are trained using mini-batch stochastic gradient descent with a momentum fixed to 0.9. Our implementation is based on the publicly available Caffe toolbox~\cite{jia2014caffe} with some modifications. The input video frame is uniformly fixed to the size of 224$\times$224.  In addition, we also perform simple data augmentations like cropping and flipping following~\cite{DBLP:conf/nips/SimonyanZ14}. 

The spatial and the audio ConvNets are first pre-trained using the ILSVRC-2012 training set with 1.2 million images and then fine-tuned using the training video data. This strategy has been observed effective in~\cite{DBLP:conf/nips/SimonyanZ14} for the spatial stream, and we have observed it also helpful for the audio stream. To fine-tune the spatial and the audio ConvNets, we gradually decrease the learning rate from $10^{-3}$ to $10^{-4}$ after 14K iterations, then to $10^{-5}$ after 20K iterations.
In addition, dropout is applied to the fully connected layers with a ratio of 0.5 to avoid over-fitting.

To train the motion ConvNet, we first compute optical flow using the GPU implementation of~\cite{brox2004high} and stack the optical flows in each 10-frame window to receive a 20-channel optical flow image as the input (one horizontal channel and one vertical channel for each frame pair). Unlike the spatial and the audio ConvNets, we train the motion ConvNet from scratch by adopting 0.7 dropout ratio and setting the learning rate to $10^{-2}$ initially, which is reduced to to $10^{-3}$ after 100K iterations and then to $10^{-4}$ after 200K iterations. Note that we also tried to use the VGG\_19 network to train the motion ConvNet, but observed worse results as the network contains much more parameters that cannot be well-tuned using the limited training video data. 

{\textbf{LSTM}}. We adopt the two-layer LSTM model proposed by Graves~\cite{graves2005framewise} for temporal modeling. Two models are trained with features extracted respectively from the first fully-connected layer of the spatial and the motion ConvNets as inputs. Each LSTM has 1,024 hidden units in the first layer and 512 hidden units in the second layer. We utilize a parallel implementation of the BPTT algorithm with a mini-batch size of 10 to train the network weights, where the learning rate and momentum are set as $10^{-4}$ and 0.9. In addition, we set the maximal training iterations to be 150K. Note that, in this paper, we focus on a multi-stream framework by utilizing the audio signal as a single stream for video classification. Further decomposing the audio track into multiple segments to extract more detailed temporal audio dynamics is feasible.

\textbf{Fusion}. As shown in Equation~\ref{eq:finalobj}, the proposed adaptive fusion strategy seeks a tradeoff between the empirical loss and the two regularization terms. We uniformly fix $\lambda_2$ to be $10^{-3}$ to encourage sparsity in the weight matrix. The parameter $\lambda_1$ is selected among $\{10^{-5}, 10^{-4}, 10^{-3}, 10^{-2}\}$ using cross-validation. 

\textbf{Discussions}. The proposed multi-stream framework has the capability of modeling video data comprehensively by adaptively fusing audio, static spatial, short-term motion and long-term temporal clues. As described above, such a framework consists of multiple separately trained deep networks. Although being feasible to jointly train the entire framework, it is complicated and computationally demanding. A recent work performing joint training of the LSTM with a ConvNet improves the results on the UCF-101 benchmark from 70.5\% (separate network training) to 71.1\%~\cite{DBLP:journals/corr/DonahueHGRVSD14}, which is not very significant. In addition, training multiple deep networks separately makes the approach more flexible, where a component may be replaced without the need of re-training the entire framework. For instance, one can utilize more discriminative ConvNet models like the GoogLeNet~\cite{Szegedy:2014tb} and deeper RNN models~\cite{DBLP:journals/corr/ChungGCB15} to replace the current ConvNet and LSTM parts respectively for better performance. Therefore, in this work, we focus on presenting a general framework for multi-stream video classification. With the proposed adaptive fusion method, such a multi-stream framework is empirically proved to be effective for the video classification task, as discussed in the following section. 

\section{Experiments}
In this section, we report results on two popular datasets. Experiments are designed to study the effectiveness of each individual stream and the proposed adaptive multi-stream fusion method. 

\subsection{Experimental Setup}
\textbf{Datasets and Evaluation Measures}. UCF-101~\cite{ucf101} is a widely adopted dataset for human action recognition, containing 13,320 video clips annotated into 101 action classes. All the video clips have a fixed frame rate of 25 fps with a spatial resolution of 320 $\times$ 240 pixels. This dataset is challenging because most videos were captured under uncontrolled environments with camera motion, cluttered backgrounds and large intra-class variations. We follow the suggested experimental protocol and report mean accuracy over the three training and test splits. 

The Columbia Consumer Videos (CCV) dataset~\cite{icmr11:consumervideo} contains 9,317 YouTube videos and 20 classes. Most of the classes are events like ``basketball'', ``graduation ceremony'' and ``wedding dance''. A few are scenes and objects like ``beach" and ``dog". Following~\cite{icmr11:consumervideo}, we adopt the suggested training and test split and compute the average precision (AP) for each class. Mean AP (mAP) is used to measure the overall performance on this dataset. 

The two datasets possess very different characteristics. Besides the difference of the defined semantic classes, the average video duration of CCV is 80 seconds, which is around ten times longer than that of UCF-101. Testing on these two datasets is helpful for evaluating the effectiveness and the generalization capability of our multi-stream classification approach.

\textbf{Alternative Fusion Methods}. To validate the effectiveness of our adaptive multi-stream fusion method, we compare with the following alternatives: (1) Average Fusion, where the mean scores of multiple networks are used as the final prediction; (2) Weighted Fusion, where the scores are fused linearly with weights estimated by cross-validation; (3) Kernel Average Fusion, where the scores are used as features and kernels computed from different network scores are averaged to train an SVM classifier; (4) Multiple Kernel Learning (MKL) Fusion, where the kernels are combined using the $\ell_p$-norm MKL algorithm~\cite{kloft2011lp}; (5) Logistic Regression Fusion, where a logistic regression model is trained to estimate the fusion weights.

\subsection{Results and Discussions}
\subsubsection{Multi-Stream Networks}
We first report the performance of each individual stream on both datasets. After that, average fusion is adopted to study whether two or more streams are complementary. The proposed adaptive fusion method will be evaluated later.

Table~\ref{tbl:stream} reports the results. Comparing the top two cells of results on UCF-101, it is interesting to observe that the spatial LSTM outperforms the spatial ConvNet and the motion LSTM is also comparable to the motion ConvNet. This is largely due to the fact that the long-term temporal clues are fully discarded in the ConvNet based classification, which contain valuable information that can be exploited by the LSTM. 

On the CCV dataset, the ConvNet achieves significantly better results than the LSTM on both spatial and motion streams. This is because the classes in CCV are either high-level events or objects/scenes. Compared with human actions, the temporal clues of these classes are more obscure and thus difficult to be captured. Also, the CCV videos are temporally untrimmed, which may contain significant portions of contents irrelevant to the classes, making the temporal modeling task even more difficult. 

The audio ConvNets operated on spectrograms produce 16.2\% on UCF-101 and 21.5\% on CCV. Note that only 51 classes in UCF-101 have audio signals, and the performance on the 51-class subset is actually 32.1\%. The audio stream is much worse than the spatial and the motion streams on both datasets, confirming that the visual channel are more informative than the audio counterpart. 

\begin{table}[t]
\begin{center}
\begin{tabular}{|c|c|c|}
\hline
                        & UCF-101         & CCV       \\ \hline   \hline 
Spatial ConvNet            & 80.4         & 75.0   \\ 
Motion ConvNet             & 78.3         & 59.1    \\ \hline \hline
Spatial LSTM            & 83.3         & 43.3   \\ 
Motion LSTM             & 76.6            & 54.7      \\ \hline \hline
Audio ConvNet           & 16.2$^*$     & 21.5   \\ \hline  \hline
ConvNet (spatial+motion)  & 86.2         & 75.8 \\
LSTM (spatial+motion)       & 86.3        & 61.9    \\ \hline \hline
ConvNet+LSTM (spatial)    & 84.0       & 77.9   \\ 
ConvNet+LSTM (motion)      & 81.4         & 70.9   \\ \hline \hline
ConvNet+LSTM (spatial+motion)    
                     & 90.1         & 81.7   \\     
All the streams               & 90.3         & 82.4   \\ \hline 
\end{tabular}
\caption{Performance of each individual stream and their average fusion (indicated by ``+"). $^*$Note that only the videos of 51 classes in UCF-101 contain audio soundtracks. The audio ConvNet can produce an accuracy of 32.1\% on the 51-class subset. }
\label{tbl:stream} 
\end{center}
\vspace{-0.2in}
\end{table}

Next, we evaluate the combinations of multiple networks to study whether fusion can compensate the limitations of a single stream in describing complex video data. The simple average fusion is adopted. Results are summarized in the bottom three groups of Table~\ref{tbl:stream}. We first assess the gain from integrating the spatial and the motion information modeled by ConvNet and LSTM respectively. On UCF-101, significant improvements (about 6\% for ConvNet and 3\% for LSTM) are observed over the best single stream results. The gain on CCV is consistent but not as significant as that on UCF-101, indicating that the short-term motion is more critical for human action analysis. Note that the average fusion of the spatial and the motion ConvNets follows the same idea of the two-stream approach proposed in~\cite{DBLP:conf/nips/SimonyanZ14}. Our implementation of this approach produces slightly worse performance than that originally reported in~\cite{DBLP:conf/nips/SimonyanZ14} (86.2\% vs. 88.0\%).

We also fuse ConvNet with LSTM separately on both streams to investigate the contribution of the long-term temporal modeling. Overall, we observe very consistent improvements on both datasets. In particular, on CCV, although the individual LSTM model is worse than ConvNet, the combination of them leads to significant  improvements. Especially, a gain of nearly 12\% is obtained on the motion stream. These results show that the long-term temporal clues are highly complementary to the ConvNet-based predictions, even in the case of modeling complex contents in the long CCV videos, which is fairly appealing.

Finally, the combination of ConvNet and LSTM on both streams, indicated by ``ConvNet+LSTM (spatial+motion)", achieves 90.1\% and 81.7\% on UCF-101 and CCV respectively. Further adding the audio ConvNet (``all the streams") can improve the results particularly on CCV which contains many classes that can be partly revealed by auditory clues (\eg, cheering sounds in the sports events). In summary, the fusion results clearly demonstrate that all the multimodal clues in our approach are useful and should be adopted in a successful video classification system.

\subsubsection{Adaptive Multi-Stream Fusion}
In this subsection, we evaluate the proposed adaptive multi-stream fusion approach, and compare it with the alternative methods. Table~\ref{tbl:fusion} gives the results. We see that all the methods produce better results than the individual streams. The simple average fusion and weighted fusion are slightly better than the learning based kernel fusion and logistic regression fusion, indicating that learning to fuse the prediction scores in a ``free'' manner is prone to over-fitting. Kernel average fusion shows slightly better results than MKL, which is consistent with the observations in several previous studies like~\cite{gehler2009feature}.

Our proposed adaptive multi-stream fusion (the bottom row) outperforms all the alternatives with clear margins. To investigate the contributions of the two regularizers in our approach, we set $\lambda_1$ and $\lambda_2$ to be zero respectively. As can be seen, the class relationship regularizer ($\lambda_2=0$) plays a more important role than the sparsity regularizer ($\lambda_1=0$). This corroborates the effectiveness of using the class relationships, which not only brings in useful contextual information but also helps prevent over-fitting. The two regularizers are complementary as the sparsity inducing norm further enhances robustness by alleviating incorrect information sharing. Note that when eliminating both regularizers, our fusion approach degenerates to the standard logistic regression fusion.

\begin{figure*}[t]
\centering
\epsfig{file=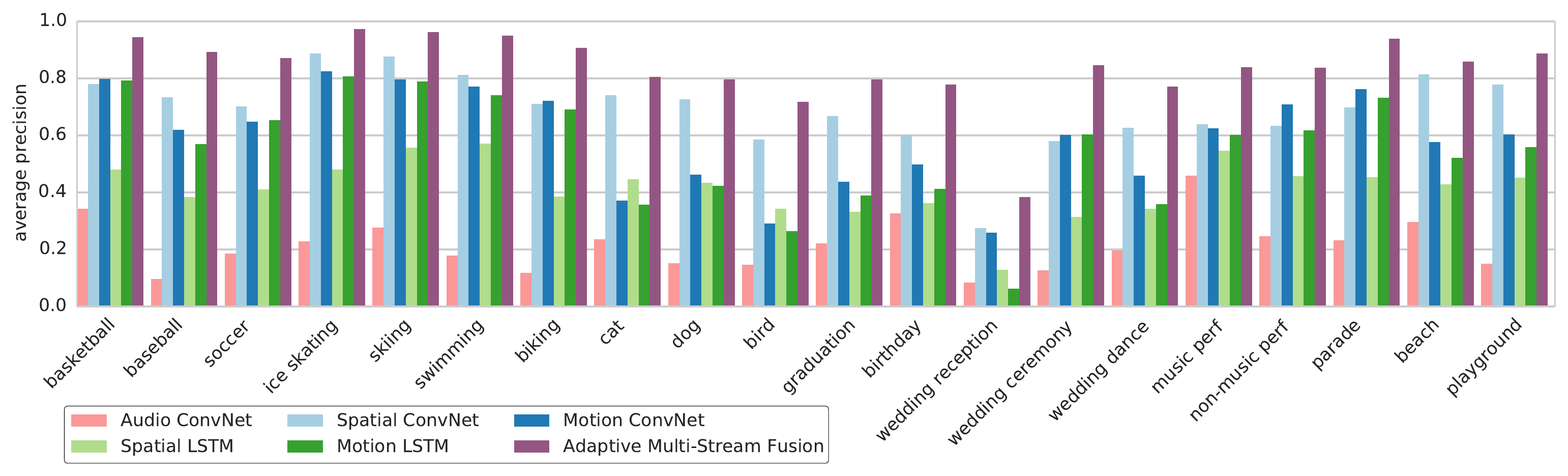, scale=0.56}
\caption{Per-class performance on CCV. Adaptive fusion of the multi-stream deep networks produces consistently better results than the individual streams on all the classes. }
\label{fig:perclassCCV}
\end{figure*}

The contribution of the audio clues is similar on both datasets (``-A" indicates the same approach without using the audio ConvNet). Audio improves just 0.4\% on UCF-101 because only half of the video clips contain soundtracks. 
Overall, the gain from the adaptive multi-stream fusion is more significant on UCF-101 as it has more classes for semantic sharing. Figure~\ref{fig:perclassCCV} further shows the per-class performance on CCV, where we can see that the fusion leads to very consistent and significant improvements for all the classes. 

\begin{table}[t]
\begin{center}
\begin{tabular}{|c|c|c|}
\hline
                                    & UCF-101         & CCV       \\ \hline   \hline 
Average fusion                         & 90.3         & 82.4    \\
Weighted fusion                     & 90.6         & 82.7    \\ 
Kernel average fusion                  & 90.2          & 82.1   \\ 
MKL fusion                                  & 89.6       & 81.8   \\ 
Logistic regression fusion                  & 89.8       & 82.0   \\ \hline \hline
Adaptive multi-stream fusion ($\lambda_1$=0)    & 90.9         & 82.8    \\ 
Adaptive multi-stream fusion ($\lambda_2$=0)    & 91.6         & 83.7   \\ 
Adaptive multi-stream fusion (-A)         & 92.2         & 84.0   \\ 
Adaptive multi-stream fusion           & 92.6         & 84.9   \\ \hline   
\end{tabular}
\caption{Comparison of fusion methods. ``-A" indicates that the audio stream ConvNet is not adopted. See texts for discussions.}
\label{tbl:fusion} 
\vspace{-0.1in}
\end{center}
\end{table}

\subsubsection{Comparison with State of the Arts}
We compare our approach with the state of the arts on both datasets.  Results are listed in Table~\ref{tbl:comparison}. Our proposed multi-stream approach achieves the highest performance on both datasets. On UCF-101, many works with competitive results are based on the hand-engineered dense trajectory features~\cite{wang2013lear,lan2014beyond}, while our approach fully relies on the deep networks. Compared with the original result of the two-stream approach~\cite{DBLP:conf/nips/SimonyanZ14}, our approach captures a more comprehensive set of useful clues with a more effective fusion strategy. Note that a gain of even just 1\% on the widely adopted UCF-101 dataset is generally considered as a significant progress.  

In addition, the recent works in~\cite{DBLP:journals/corr/DonahueHGRVSD14,DBLP:journals/corr/SrivastavaMS15,wu2015modeling,ng2015beyond} also adopted the  LSTM to model the temporal clues for video classification and reported promising performance, but did not explore the audio stream and employ advanced fusion strategies. Zha \etal~\cite{zha2015exploiting} combined the ConvNet features with the dense trajectories~\cite{wang2013action} to achieve very competitive results. 

On the CCV dataset, all the recent approaches were developed based on multiple features, either the hand-engineered descriptors or the ConvNet-based representations. Our approach produces better results than all of them.

\begin{table}[t!]
\begin{center}
\begin{tabular}{|c|c||c|c|} 
\hline
\multicolumn{2}{|c||}{UCF-101}                &    \multicolumn{2}{c|}{CCV}        \\ \hline \hline

Donahue \etal~\cite{DBLP:journals/corr/DonahueHGRVSD14}  & 82.9         & Lai \etal~\cite{lai2014video} & 43.6 \\ 

Srivastava \etal~\cite{DBLP:journals/corr/SrivastavaMS15} 
                                             & 84.3         & Jiang \etal~\cite{icmr11:consumervideo} & 59.5 \\
                                             
Wang \etal~\cite{wang2013lear}                        & 85.9         & Xu \etal~\cite{xu2013feature}    & 60.3 \\ 
Tran \etal~\cite{tran2014c3d}                         & 86.7         & Ma \etal~\cite{DBLP:journals/ijcv/MaY14}   & 63.4 \\ 
Simonyan \etal~\cite{DBLP:conf/nips/SimonyanZ14}            & 88.0         & Jhuo \etal~\cite{MVA:audiovisual} & 64.0 \\ 

Ng \etal~\cite{ng2015beyond}                       & 88.6      & Ye \etal~\cite{ye2012robust}      & 64.0 \\ 
Lan \etal~\cite{lan2014beyond}                        & 89.1         & Liu \etal~\cite{liu2013sample}    & 68.2 \\ 
Zha \etal~\cite{zha2015exploiting}                    & 89.6         & Wu \etal~\cite{wu2015modeling}   & 83.5 \\ 
Wu \etal~\cite{wu2015modeling}                        & 91.3         &           & \\ \hline \hline
Ours (-A)                                       & 92.2         & Ours (-A) & 84.0 \\ 
Ours                                            &  {\bf 92.6}  & Ours &  {\bf 84.9} \\ \hline
\end{tabular}
\caption{\label{tbl:comparison} Comparison with state-of-the-art results. Our approach produces to-date the highest reported results on both datasets. ``Ours (-A)" indicates the same framework without using the audio stream ConvNet.}
\label{tb:comparison}
\end{center}
\end{table}

\section{Conclusions}
We have presented a multi-stream framework of deep networks for video classification. The framework harnesses multimodal features that are more comprehensive than those previously adopted. Specifically, standard ConvNets are applied to audio spectrograms, visual frames and stacked optical flows to exploit the audio, spatial and short-term motion clues in videos, respectively. LSTM is further adopted on the spatial and the short-term motion features from the ConvNets for long-term temporal modeling. The outputs from the different streams are then fused using a novel method that adaptively learns the fusion weights for each class. Through imposing regularizations with the prior information and the sparsity, the weight learning process explores semantic class correlations, while suppressing inappropriate knowledge sharing among irrelevant classes. Our results confirm that all the adopted streams are effective for modeling not only simple human actions in short clips but also complex events in temporally untrimmed videos on the Internet. Combining all the streams by our proposed adaptive fusion method outperforms peer approaches with significant margins on two popular benchmarks.

The work in this paper is among the very few studies showing strong video classification performance using deep networks. As aforementioned, unlike the spatial ConvNet that can be trained by fine-tuning a model pre-trained on the ImageNet dataset, the motion ConvNet has to be trained from scratch on videos. Therefore, one promising future direction is to pre-train the motion ConvNet using large video datasets like the Sports-1M~\cite{KarpathyCVPR14}, which may improve the results significantly.

\small
\bibliographystyle{ieee}
\bibliography{reference}

\begin{thebibliography}{10}\itemsep=-1pt

\bibitem{abdel2013exploring}
O.~Abdel-Hamid, L.~Deng, and D.~Yu.
\newblock Exploring convolutional neural network structures and optimization
  techniques for speech recognition.
\newblock In {\em INTERSPEECH}, 2013.

\bibitem{AssariUCF-CVPR2014}
S.~M. Assari, A.~R. Zamir, and M.~Shah.
\newblock Video classification using semantic concept co-occurrences.
\newblock In {\em CVPR}, 2014.

\bibitem{bach2011convex}
F.~Bach, R.~Jenatton, J.~Mairal, G.~Obozinski, et~al.
\newblock Convex optimization with sparsity-inducing norms.
\newblock {\em Optimization for Machine Learning}, 2011.

\bibitem{bengio2013using}
S.~Bengio, J.~Dean, D.~Erhan, E.~Ie, Q.~Le, A.~Rabinovich, J.~Shlens, and
  Y.~Singer.
\newblock Using web co-occurrence statistics for improving image
  categorization.
\newblock {\em CoRR}, 2013.

\bibitem{bengio1994learning}
Y.~Bengio, P.~Simard, and P.~Frasconi.
\newblock Learning long-term dependencies with gradient descent is difficult.
\newblock {\em IEEE TNN}, 1994.

\bibitem{brox2004high}
T.~Brox, A.~Bruhn, N.~Papenberg, and J.~Weickert.
\newblock High accuracy optical flow estimation based on a theory for warping.
\newblock In {\em ECCV}. 2004.

\bibitem{DBLP:journals/corr/ChungGCB15}
J.~Chung, {\c{C}}.~G{\"{u}}l{\c{c}}ehre, K.~Cho, and Y.~Bengio.
\newblock Gated feedback recurrent neural networks.
\newblock {\em CoRR}, 2015.

\bibitem{DengJiaECCV2014}
J.~Deng, N.~Ding, Y.~Jia, A.~Frome, K.~Murphy, S.~Bengio, Y.~Li, H.~Neven, and
  H.~Adam.
\newblock Large-scale object classification using label relation graphs.
\newblock In {\em ECCV}, 2014.

\bibitem{DBLP:journals/corr/DonahueHGRVSD14}
J.~Donahue, L.~A. Hendricks, S.~Guadarrama, M.~Rohrbach, S.~Venugopalan,
  K.~Saenko, and T.~Darrell.
\newblock Long-term recurrent convolutional networks for visual recognition and
  description.
\newblock {\em CoRR}, 2014.

\bibitem{donoho1995adapting}
D.~L. Donoho and I.~M. Johnstone.
\newblock Adapting to unknown smoothness via wavelet shrinkage.
\newblock {\em Journal of the american statistical association}, 1995.

\bibitem{farabet2013learning}
C.~Farabet, C.~Couprie, L.~Najman, and Y.~LeCun.
\newblock Learning hierarchical features for scene labeling.
\newblock {\em IEEE TPAMI}, 2013.

\bibitem{gehler2009feature}
P.~Gehler and S.~Nowozin.
\newblock On feature combination for multiclass object classification.
\newblock In {\em ICCV}, 2009.

\bibitem{girshick2014rcnn}
R.~Girshick, J.~Donahue, T.~Darrell, and J.~Malik.
\newblock Rich feature hierarchies for accurate object detection and semantic
  segmentation.
\newblock In {\em CVPR}, 2014.

\bibitem{DBLP:conf/icassp/GravesMH13}
A.~Graves, A.~Mohamed, and G.~E. Hinton.
\newblock Speech recognition with deep recurrent neural networks.
\newblock In {\em ICASSP}, 2013.

\bibitem{graves2005framewise}
A.~Graves and J.~Schmidhuber.
\newblock Framewise phoneme classification with bidirectional lstm and other
  neural network architectures.
\newblock {\em Neural Networks}, 2005.

\bibitem{hochreiter1997long}
S.~Hochreiter and J.~Schmidhuber.
\newblock Long short-term memory.
\newblock {\em Neural computation}, 1997.

\bibitem{MVA:audiovisual}
I.-H. Jhuo, G.~Ye, S.~Gao, D.~Liu, Y.-G. Jiang, D.~T. Lee, and S.-F. Chang.
\newblock Discovering joint audio-visual codewords for video event detection.
\newblock {\em Machine Vision and Applications}, 2014.

\bibitem{DBLP:conf/icml/JiXYY10}
S.~Ji, W.~Xu, M.~Yang, and K.~Yu.
\newblock 3d convolutional neural networks for human action recognition.
\newblock In {\em ICML}, 2010.

\bibitem{jia2014caffe}
Y.~Jia, E.~Shelhamer, J.~Donahue, S.~Karayev, J.~Long, R.~Girshick,
  S.~Guadarrama, and T.~Darrell.
\newblock Caffe: Convolutional architecture for fast feature embedding.
\newblock In {\em ACM Multimedia}, 2014.

\bibitem{icmr11:consumervideo}
Y.-G. Jiang, G.~Ye, S.-F. Chang, D.~Ellis, and A.~C. Loui.
\newblock Consumer video understanding: A benchmark database and an evaluation
  of human and machine performance.
\newblock In {\em {ACM} ICMR}, 2011.

\bibitem{KarpathyCVPR14}
A.~Karpathy, G.~Toderici, S.~Shetty, T.~Leung, R.~Sukthankar, and L.~Fei-Fei.
\newblock Large-scale video classification with convolutional neural networks.
\newblock In {\em CVPR}, 2014.

\bibitem{kloft2011lp}
M.~Kloft, U.~Brefeld, S.~Sonnenburg, and A.~Zien.
\newblock Lp-norm multiple kernel learning.
\newblock {\em The Journal of Machine Learning Research}, 2011.

\bibitem{DBLP:conf/nips/KrizhevskySH12}
A.~Krizhevsky, I.~Sutskever, and G.~E. Hinton.
\newblock Imagenet classification with deep convolutional neural networks.
\newblock In {\em NIPS}, 2012.

\bibitem{lai2014video}
K.-T. Lai, F.~X. Yu, M.-S. Chen, and S.-F. Chang.
\newblock Video event detection by inferring temporal instance labels.
\newblock In {\em CVPR}, 2014.

\bibitem{lan2014beyond}
Z.~Lan, M.~Lin, X.~Li, A.~G. Hauptmann, and B.~Raj.
\newblock Beyond gaussian pyramid: Multi-skip feature stacking for action
  recognition.
\newblock {\em CoRR}, 2014.

\bibitem{laptevSTIP}
I.~Laptev.
\newblock On space-time interest points.
\newblock {\em IJCV}, 2007.

\bibitem{liu2013sample}
D.~Liu, K.-T. Lai, G.~Ye, M.-S. Chen, and S.-F. Chang.
\newblock Sample-specific late fusion for visual category recognition.
\newblock In {\em CVPR}, 2013.

\bibitem{DBLP:journals/ijcv/Lowe04}
D.~G. Lowe.
\newblock Distinctive image features from scale-invariant keypoints.
\newblock {\em IJCV}, 2004.

\bibitem{DBLP:journals/ijcv/MaY14}
A.~J. Ma and P.~C. Yuen.
\newblock Reduced analytic dependency modeling: Robust fusion for visual
  recognition.
\newblock {\em IJCV}, 2014.

\bibitem{nandakumar2008likelihood}
K.~Nandakumar, Y.~Chen, S.~C. Dass, and A.~K. Jain.
\newblock Likelihood ratio-based biometric score fusion.
\newblock {\em IEEE TPAMI}, 2008.

\bibitem{natarajan2012multimodal}
P.~Natarajan, S.~Wu, S.~Vitaladevuni, X.~Zhuang, S.~Tsakalidis, U.~Park, and
  R.~Prasad.
\newblock Multimodal feature fusion for robust event detection in web videos.
\newblock In {\em CVPR}, 2012.

\bibitem{ng2015beyond}
J.~Y.-H. Ng, M.~Hausknecht, S.~Vijayanarasimhan, O.~Vinyals, R.~Monga, and
  G.~Toderici.
\newblock Beyond short snippets: Deep networks for video classification.
\newblock {\em CoRRs}, 2015.

\bibitem{oneata2013action}
D.~Oneata, J.~Verbeek, C.~Schmid, et~al.
\newblock Action and event recognition with fisher vectors on a compact feature
  set.
\newblock In {\em ICCV}, 2013.

\bibitem{rabinovich2007objects}
A.~Rabinovich, A.~Vedaldi, C.~Galleguillos, E.~Wiewiora, and S.~Belongie.
\newblock Objects in context.
\newblock In {\em ICCV}, 2007.

\bibitem{sanchez2013image}
J.~S{\'a}nchez, F.~Perronnin, T.~Mensink, and J.~Verbeek.
\newblock Image classification with the fisher vector: Theory and practice.
\newblock {\em IJCV}, 2013.

\bibitem{DBLP:conf/nips/SimonyanZ14}
K.~Simonyan and A.~Zisserman.
\newblock Two-stream convolutional networks for action recognition in videos.
\newblock In {\em NIPS}, 2014.

\bibitem{simonyan2014very}
K.~Simonyan and A.~Zisserman.
\newblock Very deep convolutional networks for large-scale image recognition.
\newblock {\em CoRR}, 2014.

\bibitem{ucf101}
K.~Soomro, A.~R. Zamir, and M.~Shah.
\newblock {UCF101:} {A} dataset of 101 human actions classes from videos in the
  wild.
\newblock {\em CoRR}, 2012.

\bibitem{DBLP:journals/corr/SrivastavaMS15}
N.~Srivastava, E.~Mansimov, and R.~Salakhutdinov.
\newblock Unsupervised learning of video representations using {LSTMs}.
\newblock {\em CoRR}, 2015.

\bibitem{sun2009action}
X.~Sun, M.~Chen, and A.~Hauptmann.
\newblock Action recognition via local descriptors and holistic features.
\newblock In {\em CVPR}, 2009.

\bibitem{Szegedy:2014tb}
C.~Szegedy, W.~Liu, Y.~Jia, P.~Sermanet, S.~Reed, D.~Anguelov, D.~Erhan,
  V.~Vanhoucke, and A.~Rabinovich.
\newblock {Going Deeper with Convolutions}.
\newblock {\em CoRR}, 2014.

\bibitem{tran2014c3d}
D.~Tran, L.~Bourdev, R.~Fergus, L.~Torresani, and M.~Paluri.
\newblock C3d: Generic features for video analysis.
\newblock {\em CoRR}, 2014.

\bibitem{van2013deep}
A.~Van~den Oord, S.~Dieleman, and B.~Schrauwen.
\newblock Deep content-based music recommendation.
\newblock In {\em NIPS}, 2013.

\bibitem{wang2013action}
H.~Wang and C.~Schmid.
\newblock Action recognition with improved trajectories.
\newblock In {\em ICCV}, 2013.

\bibitem{wang2013lear}
H.~Wang and C.~Schmid.
\newblock Lear-inria submission for the thumos workshop.
\newblock {\em ICCV THUMOS Workshop}, 2013.

\bibitem{wu2015modeling}
Z.~Wu, X.~Wang, Y.-G. Jiang, H.~Ye, and X.~Xue.
\newblock Modeling spatial-temporal clues in a hybrid deep learning framework
  for video classification.
\newblock {\em CoRR}, 2015.

\bibitem{xu2013feature}
Z.~Xu, Y.~Yang, I.~Tsang, N.~Sebe, and A.~Hauptmann.
\newblock Feature weighting via optimal thresholding for video analysis.
\newblock In {\em ICCV}, 2013.

\bibitem{yao2015video}
L.~Yao, A.~Torabi, K.~Cho, N.~Ballas, C.~Pal, H.~Larochelle, and A.~Courville.
\newblock Video description generation incorporating spatio-temporal features
  and a soft-attention mechanism.
\newblock {\em CoRR}, 2015.

\bibitem{ye2012robust}
G.~Ye, D.~Liu, I.-H. Jhuo, and S.-F. Chang.
\newblock Robust late fusion with rank minimization.
\newblock In {\em CVPR}, 2012.

\bibitem{zha2015exploiting}
S.~Zha, F.~Luisier, W.~Andrews, N.~Srivastava, and R.~Salakhutdinov.
\newblock Exploiting image-trained cnn architectures for unconstrained video
  classification.
\newblock {\em CoRR}, 2015.

\end{thebibliography}
\end{document}